\documentclass[10pt, a4paper]{article}
\usepackage{lrec2016}
\usepackage{multibib}
\newcites{languageresource}{Language Resources}
\usepackage{graphicx}
\usepackage{epstopdf}
\usepackage[latin1]{inputenc}
\usepackage{xurl}

\title{POS tagging, lemmatization and dependency parsing of West Frisian\\ }

\name{Wilbert Heeringa$^1$, Gosse Bouma$^2$, Martha Hofman$^1$, Eduard Drenth$^1$,\vspace{0.1cm}\\
      \textbf{\large Jan Wijffels$^3$, Hans Van de Velde$^{1, 4}$}\vspace{0.3cm}}

\address{$^1$Fryske Akademy, $^2$University of Groningen, $^3$BNOSAC, $^4$Utrecht University\\
         $^1$Leeuwarden, $^2$Groningen, $^3$Brussels, $^4$Utrecht\\
         $^1$\texttt{\{}wheeringa, mhofman, edrenth, hvandevelde\texttt{\}}@fryske-akademy.nl,\\
         $^2$g.bouma@rug.nl, $^3$jwijffels@bnosac.be\\}

\abstract{
We present a lemmatizer/POS-tagger/dependency parser for West Frisian using a corpus of 44,714 words in 3,126 sentences that were annotated according to the guidelines of Universal Dependency version 2. POS tags were assigned to words by using a Dutch POS tagger that was applied to a literal word-by-word translation, or to sentences of a Dutch parallel text. Best results were obtained when using literal translations that were created by using the Frisian translation program \textit{Oersetter}. Morphologic and syntactic annotations were generated on the basis of a literal Dutch translation as well. The performance of the lemmatizer/tagger/annotator when it was trained using default parameters was compared to the performance that was obtained when using the parameter values that were used for training the LassySmall UD 2.5 corpus. A significant improvement was found for `lemma'. The Frisian lemmatizer/PoS tagger/dependency parser is released as a web app and as a web service.\\ 
\newline 
\Keywords{keyword A, keyword B, keyword C} }

\begin{document}

\maketitleabstract

\section{Introduction}

Since small languages are struggling with restricted possibilities for developing language tools, there is a need to develop an efficient strategy that makes the development of tools feasible, for example by taking a route via a larger and related language for which the desired tools have already been developed. There are several papers that focus on tagging (historical) texts via a closely-related language. 

In the simplest case texts of a small language are tagged by taggers which are trained on the basis of tagged corpora of closely-related languages. This approach is mentioned by \newcite{scherrer:2014} who tagged Catalan data using a Spanish PoS-tagger. He obtained an accuracy of 58.42\%. \newcite{tjongkimsang:2016} tagged 17th-century Dutch texts with a modern Dutch POS-tagger and got accuracies of 62.8\% and 63.7\%. The low accuracies show that this approach is not enough, but it can serve a baseline to which other approaches can be compared. In the case of West Frisian, we consider using a tagger that is trained on the basis of Dutch texts.

\newcite{scherrer:2014} obtained higher accuracies of up to 89.08\% when the Spanish tagger was adapted to Catalan. Words in the tagger parameter files were translated into the low-resource language by using translation dictionaries that were created with unsupervised lexicon induction techniques that rely only on raw textual data.

Results can also be improved by adapting the spelling of corpora prior to tagging them with a tagger. For example, \newcite{hupkes:2016} investigated PoS-tagging of 17th-century Dutch texts. They found that ``that modernizing the spelling of corpora prior to tagging them with a tagger trained on contemporary Dutch results in a large increase in accuracy, but that spelling normalization alone is not sufficient to obtain state-of-the-art results.''

\newcite{yarowsky:2001} initiated an approach that draws on parallel corpora. The source side of a parallel corpus is tagged with an existing tagger and the tags along the word alignment links are projected on the target side. Then a new tagger is trained on the target side.

\newcite{tjongkimsang:2016} PoS-tagged 17th-century Dutch texts. He translated the texts into modern Dutch word by word. On the basis of the modern Dutch translation the texts were PoS tagged with Frog. Compared the the previously mentioned method the advantage is that no parallel text is required. \newcite{tjongkimsang:2016} evaluated four word-by-word translation methods. In the first method, the machine translation system Moses was used with two versions of the Dutch Statenvertaling Bible, one from the year 1637 and one from 1888. The second approach used the Integrated Language Bank (GTB), an online collection of historical dictionaries, with links of historical words to their modern counterparts. The lexicon service makes it possible to retrieve modern lemmas for historical words. The third used a lexicon that was learned from two versions of a Dutch Bible, one from the seventeenth century and one from the nineteenth century. The fourth method employed orthographic rules learned from the learned lexicon. The rules converted historical character sequences to their modern equivalent. The lexicon-based methods outperformed the method that used orthographic rules.

In this paper we present a corpus of lemmatized, tagged and annotated text in the West Frisian language together with a web application and web service which can be used for lemmatization, part-of-speech (PoS) tagging and dependency parsing of Frisian text. West Frisian is an autochthonous minority language spoken in the Dutch province of Frysl{\^a}n where it is recognized as an official language, in addition to Dutch. In 2018 about 400,000 native speakers were reported \cite{klinkenberg:2018}.

The corpus was built by taking a route via Standard Dutch. We investigated some of the techniques mentioned above. In Section~\ref{buildingthecorpus} the building of the corpus is described, and different procedures are compared to each other. In Section~\ref{trainingandperformance} the training of the lemmatizer/tagger/annotator is described and performance measurements are provided. In Section~\ref{toolsandresults} a web app and a web service are presented and some results are shown. Conclusions and future prospects are given in Section~\ref{conclusionsandfutureprospects}.


\section{Building the corpus}
\label{buildingthecorpus}

\subsection{Source texts}

We aimed to develop a Frisian lemmatizer/PoS tagger/dependency parser that can be used for a wide range of texts. Therefore, texts of different genres should be included in the corpus. We considered several (well-known) corpora that were PoS tagged: the Brown Corpus, the Lancaster-Oslo-Bergen Corpus, the Stockholm-Ume{\aa} Corpus, the Balanced Corpus of Contemporary Written Japanese, the NOAH's Corpus of Swiss German Dialects, the STEVIN reference corpus of 500 million words, the SoNaR corpus. In all of these corpora we found news texts. Novels, legal texts/laws and Wikipedia texts were found in three corpora, and scientific texts were found in two corpora. We decided to include texts of the same genres in our newly built corpus.

Texts were selected from the Frisan-Dutch parallel corpus that is used for the online Frisan/Dutch - Dutch/Frisian translation program \textit{Oersetter} `translator'. The corpus consists of 120 texts that are available both in Frisian and Dutch. The Frisian texts consists of 3,002,327 words, and the Dutch texts consists of 5,057,248 words.

We added texts from the Frisian broadcast station \textit{Omrop Frysl{\^a}n} as well as the Wikipedia text `Frysk'. The distributution of the genres in the newly compiled corpus is shown in Table~\ref{corpus}. Legal texts are not included yet.

\begin{table*}[ht]
  \begin{center}
    \begin{tabular}{lrrrrrr}                                                              \hline
       source    & \#tokens & \%tokens & \#words & \%words & \#sentences & \%sentences \\ \hline
       news      &     8737 &       17 &    7998 &      17 &         582 &          19 \\
       science   &     2293 &        4 &    2069 &       5 &         107 &           3 \\
       novels    &    17176 &       34 &   14272 &      32 &        1446 &          46 \\
       museum    &     9275 &       18 &    8335 &      19 &         486 &          16 \\
       Wikipedia &    13780 &       27 &   12040 &      27 &         505 &          16 \\ \hline
       total     &    51261 &          &   44714 &         &        3126 &             \\ \hline
    \end{tabular}
    \caption{Distribution of genres across the newly compiled corpus.}
    \label{corpus}
  \end{center}
\end{table*}

\subsection{Universal dependencies}

We decided to annotate the corpus according to the guidelines of Universal Dependencies (UD) version 2\footnote{See \url{https://universaldependencies.org/guidelines.html}.}. UD is a project that aims to develop a cross-linguistically consistent treebank annotation for many languages. ``The general philosophy is to provide a universal inventory of categories and guidelines to facilitate consistent annotation of similar constructions across languages, while allowing language-specific extensions when necessary.''\footnote{See \url{https://universaldependencies.org/introduction.html}.}

\subsection{Adding lemmas}

In the corpus lemmas of the tokens were added manually. When the corpus consisted of about 30.000 tokens, a lemmatizer was trained on the basis of the lemmatized Frisian texts using the function \texttt{udpipe{\textunderscore}train} from the R package \texttt{udpipe} \cite{wijffels:2020}. Using this tagger, newly added corpus texts were lemmatized, and the lemmas were manually corrected. The techniques that are made available in the R package \texttt{udpipe} are described by \newcite{straka:2016} and \newcite{straka:2017}.

\subsection{Adding PoS tags}

For adding PoS tags three procedures were investigated. In the first procedure the Frisian text was tagged with a Dutch PoS-tagger that was trained with the UD Dutch LassySmall treebank. This corpus contains sentences from the Wikipedia section of the Lassy Small Treebank. The Lassy Small Treebank is a manually verified treebank for Dutch. Since not all material in the Lassy Small Treebank could be made freely available, only the material from the Wikipedia (wiki) section was included in the UD Dutch LassySmall treebank \cite{vannoord:2013}.

In the second procedure we followed \newcite{tjongkimsang:2016}. A Dutch literal word-for-word translation is made from the the Frisian text. This literal translation is tagged by the Dutch PoS tagger that was trained with the \mbox{Lassysmall UD 2.8} corpus.\footnote{The corpus is found at \url{https://github.com/Universal Dependencies/UD_Dutch-LassySmall.}} The tags of the Dutch words are projected on the corresponding Frisian words. In order to obtain a literal translation, we investigated three translation programs:

\begin{itemize}
    \item Google Translate. Google Translate uses a neural machine translation engine called \textit{Google Neural Machine Translation} (GNMT). With GNMT the quality of the tranlation is improved by applying an example-based machine translation method in which the system ``learns from millions of examples'' \cite{schuster:2016}. We called the API by using the function \texttt{gl{\textunderscore}translate} from the R package \texttt{googleLanguageR} \cite{edmondson:2020}. As \newcite{turovsky:2016} wrote, GNMT translates ``whole sentences at a time, rather than just piece by piece...'' This makes Google Translate less suitable for generating a word-for-word translation. The best way to translate an individual word is submitting the word being enclosed in single quotes.

    \item The online Frisan/Dutch - Dutch/Frisian translation program \textit{Oersetter} `translator'\footnote{Available at \url{https://taalweb.frl/oersetter}.}. This program is a statistical machine translation (SMT) system. A parallel training corpus has been compiled and been used to automatically learn a phrase-based SMT model\cite{vangompel:2014}. The translation system is built around the open-source SMT software Moses. We used the web service of the \textit{Oersetter}.

    \item A newer version of the \textit{Oersetter}\footnote{Available at \url{https://frisian.eu/TEST/oersetter/}.}. This program is a neural machine translation (NMT) system. More specifically a transformer-based sequence-to-sequence model is employed. Marian-NMT is used as the sofware that powers the program. With the new \textit{Oersetter} higher scores on automatic evaluation measures were achieved.\footnote{See experiments 5 and 6 at \url{https://bitbucket.org/fryske-akademy/oersetter2-pipeline/src/master/data/evaluation/README.md}.} Like Google Translate this program is intended to translate word phrases or sentences. In order to get an individual word translated by an individual word, the word to be translated should be put between double quotes. We used the web service of this program\footnote{We used the version that was online on June 3rd, 2021.}
\end{itemize}

In the third procedure the approach that was initiated by \newcite{yarowsky:2001} was used. The Dutch parallel text was tagged with the tagger that was trained with the LassySmall corpus. Then the sentences of the Dutch texts were aligned with the corresponding sentences in the Frisian texts using \texttt{fast{\textunderscore}align} \cite{dyer:2013}\footnote{See
\url{https://github.com/clab/fast_align}.} and the tags of the words in the Dutch sentences were projected on the corresponding Frisian words in the Frisian sentences.

We tested the different procedures to the scientific texts, the museum texts, a part of the news texts and a part of the novel texts. The results are shown in Table~\ref{translation}. The highest percentage was found when tags were obtained via a literal Dutch translation that was obtained with the (old) Oersetter. 

\begin{table}[ht]
  \begin{center}
    \begin{tabular}{lr}                          \hline
                                 & \% correct \\ \hline
       Baseline                  &     51.5\% \\
       Alignment with Dutch text &     74.2\% \\
       Google Translate          &     76.0\% \\
       Oersetter                 &     89.8\% \\
       Oersetter new             &     81.2\% \\ \hline
    \end{tabular}
    \caption{Percentages of correct PoS tags per translation procedure.}
    \label{translation}
  \end{center}
\end{table}

\begin{table}[ht]
  \begin{center}
    \begin{tabular}{lcll}                                   \hline
                      &     &                & $p$ value \\ \hline
       Align. Dutch   & $>$ & Baseline       & $< .0001$ \\
       Google Transl. & $>$ & Align. Dutch   & $< .001$  \\
       Oersetter new  & $>$ & Google Transl. & $< .0001$ \\
       Oersetter      & $>$ & Oersetter new  & $< .0001$ \\ \hline
    \end{tabular}
    \caption{Comparison of percentages of correct PoS tags among translation procedures.}
    \label{comparison}
  \end{center}
\end{table}

Using Fishers exact test the percentages of correct PoS tags among the procedures were compared. The results are presented in Table~\ref{comparison}. From this table it can be concluded that the Oersetter performs significantly better than any of of the other translation procedures. The lower percentage for the technique where Frisian and Dutch sentences are aligned to each other can be explained by that fact that the aligner was not able to process swaps correctly.

By the time that we built the corpus, we started tagging the corpus via the Dutch text that was aligned to the Frisian text. Later on we tagged the corpus via literal translations that were generated by the Oersetter program. The tags were corrected manually. When the corpus consisted of about 30.000 tokens, we were able to train a Frisian tagger that had a moderate accuracy. We used the function \texttt{udpipe{\textunderscore}train} from the R package \texttt{udpipe}. Newly added Frisian texts were tagged with this tagger and manually corrected afterwards.

\subsection{Adding morphologic and syntactic annotations}

Initially, 575 Dutch sentences from Oersetter parallel corpus were morphologically/syntactically annotated with a Dutch UD annotator, and the annotations were projected on the corresponding Frisian sentences. Corrections of the annotations were made where alignments between Dutch and Frisian sentences went wrong. Later, literal Dutch word-by-word translations were generated by the Oersetter and subsequently annotated. Before annotating, the translations were manually corrected. A Dutch annotator was used that was trained on the LassySmall corpus using the function \texttt{udpipe{\textunderscore}train} from the R package \texttt{udpipe}.

Since correction work is involved in both procedures, it doesn't really matter which one is used. But the latter method also works if there is no Dutch translation of the Frisian text available. 

\section{Training and performance}
\label{trainingandperformance}

We randomized the order of the sentences in the lemmatized/tagged/annotated data set and split the sentences in training data (80\% of the sentences), validation data (10\%) and test data (10\%). 

A lemmatizer, PoS tagger and dependency parser was trained by using the function \texttt{udpipe{\textunderscore}train} in the R package \texttt{udpipe} \cite{wijffels:2020} \footnote{See also \url{https://cran.r-project.org/web/packages/udpipe/vignettes/udpipe-annotation.html}.} on the basis of the training data. As options for the tokenizer, PoS tagger, lemmatizer and dependency parser the default values were used.\footnote{See Sections 3.3, 3.4 and 3.5 at \url{https://ufal.mff.cuni.cz/udpipe/1/users-manual}.}

We present the tagger, lemmatizer and parser performance, measured on the testing portion of the data, evaluated in three different settings: using raw text only, using gold tokenization only, and using gold tokenization plus gold morphology (UPOS, XPOS, FEATS and Lemma). A $k$-fold cross-validation was carried out with $k$=10. First, the data set was split into $k$ sets. Then $k$ times (or folds) a set is selected as test set and another set is selected as validation set. When the sets are numbered from 1 to $k$ in the $i$th fold set $i$ is selected as test set, and set $i + 1$ is selected as validation set. If $i$ is $k$ the first set is selected as validation set. The joint remaining $k - 2$ sets are used as training sets. The results are shown in Table~\ref{modelperformance1} and show the mean and standard deviation across the 10 folds per metric and for each of the three settings.

\begin{table}[ht]
  \begin{center}
    \scriptsize
    \begin{tabular}{l|rr|rr|rr}                                 \hline
               & \multicolumn{2}{c|}{Raw text}
               & \multicolumn{2}{c|}{Gold tok}
               & \multicolumn{2}{c }{Gold tok + mor}    \\ \hline

               & mean  & sd   & mean  & sd    & mean & sd    \\ \hline
      f1 words & 100.0 & 0.01 &       &       &      &       \\
      f1 sents &  91.4 & 1.88 &       &       &      &       \\
      UPOS     &  94.4 & 0.48 &  94.4 &  0.47 &      &       \\
      XPOS     &  87.8 & 0.56 &  87.8 &  0.56 &      &       \\
      UFeats   &  89.6 & 0.43 &  89.6 &  0.42 &      &       \\
      AllTags  &  85.3 & 0.73 &  85.3 &  0.73 &      &       \\
      Lemma    &  94.9 & 0.48 &  94.9 &  0.48 &      &       \\
      UAS      &  72.3 & 1.32 &  72.7 &  1.26 & 78.4 & 1.18  \\
      LAS      &  66.3 & 1.35 &  66.6 &  1.32 & 73.9 & 1.36  \\ \hline
    \end{tabular}
    \caption{Lemmatizer, tagger and parser performance, measured on the testing portion of the data, evaluated in three different settings. The figures represent percentages. When training the default UDPipe settings were used.}
    \label{modelperformance1}
  \end{center}
\end{table}

Additionally, a lemmatizer, PoS tagger and dependency parser was trained using the options that were used for training on the basis of the LassySmall UD 2.5 corpus.\footnote{For the settings see \url{https://github.com/bnosac/udpipe/tree/ master/inst/models-ud-2.5 or https://lindat.mff.cuni.cz/repository/xmlui/handle/11234/1-3131} where the parameters can be found in `udpipe-ud-2.5-191206-reproducible{\textunderscore}training.zip'.} The results are shown in Table~\ref{modelperformance2}. Compared to the results in Table~\ref{modelperformance1}, we found a significant improvement for `lemma' ($p < 0.0001$)\footnote{Rand Wilcox's function \texttt{medmcp} from the Rallfun-v38 package was used. This function compares medians using a percentile bootstrap method. This function is very robust, even for small samples with a non-normal distribution \cite{wilcox:2021}.} both for `Raw text' and `Gold tok'.

\begin{table}[ht]
  \begin{center}
    \scriptsize
    \begin{tabular}{l|rr|rr|rr}                                 \hline
               & \multicolumn{2}{c|}{Raw text}
               & \multicolumn{2}{c|}{Gold tok}
               & \multicolumn{2}{c }{Gold tok + mor}         \\ \hline

               & mean  & sd   & mean  & sd    & mean & sd    \\ \hline
      f1 words & 100.0 & 0.01 &       &       &      &       \\
      f1 sents &  89.7 & 2.41 &       &       &      &       \\
      UPOS     &  94.6 & 0.28 &  94.6 &  0.30 &      &       \\
      XPOS     &  88.1 & 0.50 &  88.1 &  0.52 &      &       \\
      UFeats   &  89.8 & 0.47 &  89.8 &  0.50 &      &       \\
      AllTags  &  85.5 & 0.43 &  85.5 &  0.45 &      &       \\
      Lemma    &  96.0 & 0.25 &  96.0 &  0.25 &      &       \\
      UAS      &  72.5 & 1.15 &  73.1 &  1.12 & 78.5 & 0.87  \\
      LAS      &  66.4 & 1.08 &  67.0 &  1.13 & 73.9 & 1.21  \\ \hline
    \end{tabular}
    \caption{Lemmatizer, tagger and parser performance, measured on the testing portion of the data, evaluated in three different settings. The figures represent percentages. The settings for training on the basis of UD 2.5 corpora were used.}
    \label{modelperformance2}
  \end{center}
\end{table}

UAS (Unlabelled Attachment Score) and LAS (Labelled Attachment Score) are standard metrics to evaluate dependency parsing. UAS is the proportion of tokens whose head has been correctly assigned, and LAS is the proportion of tokens whose head has been correctly assigned with the right dependency label (subject, object, etc).

A lemmatizer, PoS tagger and dependency parser was trained using the LassySmall UD 2.8 corpus with the options that were used for training on the basis of the LassySmall UD 2.5 corpus. The LassySmall UD 2.8 corpus as it is available at GitHub is much larger than the current Frisian corpus and consists of 7341 sentences, 83571 words and 98242 tokens when we put the train, development, and test sections together. The performance results are shown in Table~\ref{modelperformance3}. 

We compared the results with the performance results in Table~\ref{modelperformance2}. For all metrics we found significant differences ($p < 0.0001$), usually the LassySmall figures are significantly better than the ones for the Frisian corpus. Exceptions are `f1 word', `f1 sents' and `lemma' (both for `Raw text' and `Gold tok').

\begin{table}[ht]
  \begin{center}
    \scriptsize
    \begin{tabular}{l|rr|rr|rr}                                 \hline
               & \multicolumn{2}{c|}{Raw text}
               & \multicolumn{2}{c|}{Gold tok}
               & \multicolumn{2}{c }{Gold tok + mor}         \\ \hline

               & mean  & sd   & mean  & sd    & mean & sd    \\ \hline
      f1 words &  99.9 & 0.03 &       &       &      &       \\
      f1 sents &  81.3 & 1.31 &       &       &      &       \\
      UPOS     &  95.6 & 0.30 &  95.9 &  0.30 &      &       \\
      XPOS     &  93.7 & 0.38 &  94.1 &  0.36 &      &       \\
      UFeats   &  95.1 & 0.26 &  95.6 &  0.26 &      &       \\
      AllTags  &  93.1 & 0.38 &  93.5 &  0.36 &      &       \\
      Lemma    &  94.2 & 0.25 &  94.4 &  0.25 &      &       \\
      UAS      &  81.3 & 0.74 &  83.4 &  0.71 & 87.1 & 0.59  \\
      LAS      &  77.5 & 0.73 &  79.4 &  0.65 & 84.0 & 0.66  \\ \hline
    \end{tabular}
    \caption{Lemmatizer, tagger and parser performance using the LassySmall UD 2.8 corpus, measured on the testing portion of the data, evaluated in three different settings. The figures represent percentages. The settings for training on the basis of UD 2.5 corpora were used.}
    \label{modelperformance3}
  \end{center}
\end{table}

\section{Tools and results}
\label{toolsandresults}

The Frisian lemmatizer/PoS tagger/dependency parser is released as a web app for human end users\footnote{See \url{https://frisian.eu/udpipeapp/}.}, as well as a web service for software to interact with\footnote{See \url{https://frisian.eu/udpipeservice/}.}. The source code is available at Bitbucket.\footnote{Online at \url{https://bitbucket.org/fryske-akademy/udpipe/src/master/}.} The user can choose from different output formats: a tab-separated file, an Excel file or a file in CoNLL-U format.

\begin{figure}[ht]
  \begin{center}
    \includegraphics[scale=0.63]{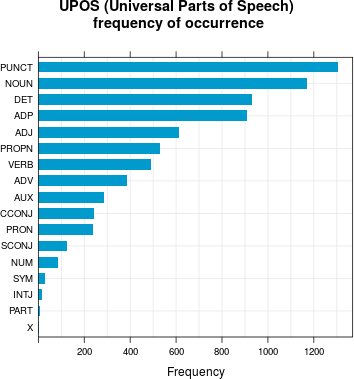} 
    \caption{Frequencies of PoS tags in the Wikipedia text `Ingelsk'.}
    \label{upos_freq}
  \end{center}
\end{figure}

\begin{figure}[ht]
  \begin{center}
    \frame{\includegraphics[scale=0.45]{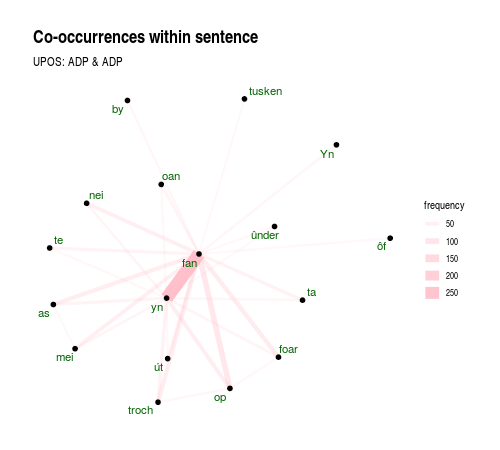}}
    \caption{Co-occurrences of prepositions within Frisian sentences in the Wikipedia text `Ingelsk'.}
    \label{adp_adp}
  \end{center}
\end{figure}

The online app has also a tab called `graphs'. The graphs provided in this tab visualize the frequencies of POS tags in the submitted text, the most frequently occurring tokens per PoS tag, frequencies of keyword word sequences, co-occurrence networks and word clouds. The graphs are made by using several R packages.\footnote{The following packages are used: \texttt{lattice}, \texttt{udpipe}, \texttt{textrank}, \texttt{igraph}, \texttt{ggraph}, \texttt{ggplot2}, \texttt{ggwordcloud}.} An example is shown in Figure~\ref{upos_freq}. The graph is obtained on the basis of the Wikipedia text 'Ingelsk' (English)\footnote{The text was retrieved on June 3rd, 2021.} and shows the frequencies of the PoS tags in the text. On the basis of  the same text Figure~\ref{adp_adp} was made. This figure visualizes frequencies of co-occurrences of prepositions within a sentence. The co-occurrence of the prepositions \textit{fan} `of' and \textit{yn} within a sentence is found to be most frequent.

\section{Conclusions and future prospects}
\label{conclusionsandfutureprospects}

In this paper we described how we built a West Frisian corpus as training material for a UD lemmatizer, PoS tagger and dependency parser. Much time is saved by not building the corpus from scratch, but rather by taking a route via a larger and related language, in this case Standard Dutch.

A Dutch tagger/annotator is applied to a (literal) Dutch translation of  the Frisian text. As for obtaining PoS tags, we investigated several alternatives such as aligning the Frisian text to a Dutch parallel text, or translating the Frisian text into Dutch by using Google Translate. However, the better PoS tags were obtained on the basis of a literal Dutch translation that was created by using the Frisian/Dutch - Dutch/Frisian translation program \textit{Oersetter}. Manual correction of the PoS tags is still necessary.

Morphological and syntactical annotations are also preferably generated on the basis of an \textit{Oersetter} translation, but the translations need to be carefully checked and corrected before submitting them to a Dutch annotator.

The current corpus consists of 44,714 words. When training a lemmatizer/PoS tagger/dependency parser on the basis of this corpus, reasonable accuracies were obtained. 

The performance of the lemmatizer/tagger/annotator when it was trained using default parameters was compared to the performance that was obtained when using the parameter values that were used for training the LassySmall UD 2.5 corpus. A significant improvement was found for `lemma'. Once the corpus is complete, the parameters will be tuned to achieve optimum performance.

We compared the performance obtained on the basis of the latter parameters with the performance of a lemmatizer/tagger/annotator that was trained on the basis of the LassySmall UD 2.8 corpus using the same parameters. The LassySmall UD 2.8 corpus is much larger than the current Frisian corpus and the performance when using this corpus was significantly better for most metrics compared to the Frisian corpus. Therefore, we aim to extend the Frisian corpus up to 100,000 words. Not only texts of the genres that are currently in the corpus will be added, but also legal texts, minutes and texts from manuals. We noticed that many mistakes are made when processing text that contains hollandisms and non-standard Frisian variants. This will also be taken into account when extending the corpus. 

The tools that we developed will be of great help for studying the Frisian language on the basis of (large) corpora. One can think of named entity resolution, coreference resolution, sentiment analysis, question answering, textometry, authorship discrimination, language detection, and measuring the influence of Dutch on Frisian.

In addition to the output formats currently provided, the XML/TEI format will be added. This will make the output also searchable for search engines like BlackLab \cite{dedoes:2017} which is used in the publicly available search system of Frisian corpora.\footnote{See \url{https://frisian.eu/frisian-corpora}, where currently mainly Middle Frisian and minimally annotated modern Frisian texts can be searched.}

\section{Acknowledgements}

This research was made possible by a CLARIAH-Plus project financed by the Dutch Research Council (Grant 184.034.023).

\newpage

\section{References}

\bibliographystyle{lrec2016}
\bibliography{main}

\begin{thebibliography}{}

\bibitem[\protect\citename{De~Does \bgroup et al.\egroup }2017]{dedoes:2017}
De~Does, J., Niestadt, J., and Depuydt, K.
\newblock (2017).
\newblock Creating research environments with {B}lack{L}ab.
\newblock In Odijk J. et~al., editors, {\em {CLARIN} in the {L}ow {C}ountries},
  pages 245--257. Ubiquity Press, London.

\bibitem[\protect\citename{Dyer \bgroup et al.\egroup }2013]{dyer:2013}
Dyer, C., Chahuneau, V., and Smith, N.~A.
\newblock (2013).
\newblock A simple, fast, and effective reparameterization of ibm model 2.
\newblock In {\em Proceedings of the 2013 Conference of the North American
  Chapter of the Association for Computational Linguistics: Human Language
  Technologies}, pages 644--648.

\bibitem[\protect\citename{Edmondson}2020]{edmondson:2020}
Edmondson, M., (2020).
\newblock {\em googleLanguageR: Call Google's 'Natural Language' API, 'Cloud
  Translation' API, 'Cloud Speech' API and 'Cloud Text-to-Speech' API}.
\newblock R package version 0.3.0.

\bibitem[\protect\citename{Hupkes and Bod}2016]{hupkes:2016}
Hupkes, D. and Bod, R.
\newblock (2016).
\newblock Pos-tagging of historical dutch.
\newblock In {\em Proceedings of the Tenth International Conference on Language
  Resources and Evaluation (LREC'16)}, pages 77--82.

\bibitem[\protect\citename{Klinkenberg \bgroup et al.\egroup
  }2018]{klinkenberg:2018}
Klinkenberg, E., Jonkman, R., and Stefan, N.
\newblock (2018).
\newblock {\em Taal yn Frysl{\^a}n. De folgjende generaasje [Language in
  Frysl{\^a}n. The next generation]}.
\newblock Morgan Kaufman Publishers, Ljouwert.

\bibitem[\protect\citename{Scherrer}2014]{scherrer:2014}
Scherrer, Y.
\newblock (2014).
\newblock Unsupervised adaptation of supervised part-of-speech taggers for
  closely related languages.
\newblock In {\em Proceedings of the First Workshop on Applying NLP Tools to
  Similar Languages, Varieties and Dialects (VarDial)}, pages 30--38.
  Association for Computational Linguistics and Dublin City University.

\bibitem[\protect\citename{Schuster \bgroup et al.\egroup }2016]{schuster:2016}
Schuster, M., Johnson, M., and Thorat, N.
\newblock (2016).
\newblock Zero-shot translation with google's multilingual neural machine
  translation system.
\newblock {\em Google AI Blog}, 22.

\bibitem[\protect\citename{Straka and Strakov{\'a}}2017]{straka:2017}
Straka, M. and Strakov{\'a}, J.
\newblock (2017).
\newblock Tokenizing, {POS} tagging, lemmatizing and parsing {UD} 2.0 with
  {UDP}ipe.
\newblock In {\em Proceedings of the {C}o{NLL} 2017 Shared Task: Multilingual
  Parsing from Raw Text to Universal Dependencies}, pages 88--99, Vancouver,
  Canada. Association for Computational Linguistics.

\bibitem[\protect\citename{Straka \bgroup et al.\egroup }2016]{straka:2016}
Straka, M., Hajic, J., and Strakov{\'a}, J.
\newblock (2016).
\newblock Udpipe: trainable pipeline for processing conll-u files performing
  tokenization, morphological analysis, pos tagging and parsing.
\newblock In {\em Proceedings of the Tenth International Conference on Language
  Resources and Evaluation (LREC'16)}, pages 4290--4297.

\bibitem[\protect\citename{Tjong Kim~Sang}2016]{tjongkimsang:2016}
Tjong Kim~Sang, E.
\newblock (2016).
\newblock Improving part-of-speech tagging of historical text by first
  translating to modern text.
\newblock In {\em International Workshop on Computational History and
  Data-Driven Humanities}, pages 54--64. Springer.

\bibitem[\protect\citename{Turovsky}2016]{turovsky:2016}
Turovsky, B.
\newblock (2016).
\newblock Found in translation: More accurate, fluent sentences in google
  translate.
\newblock {\em Blog. Google. November}, 15.

\bibitem[\protect\citename{Van~Gompel \bgroup et al.\egroup
  }2014]{vangompel:2014}
Van~Gompel, M., Van~den Bosch, A., and Dijkstra, A.
\newblock (2014).
\newblock Oersetter: Frisian-{D}utch statistical machine translation.
\newblock In P.~Boersma, et~al., editors, {\em Philologia {F}risica anno 2012},
  pages 287--296. Fryske Akademy, Ljouwert.

\bibitem[\protect\citename{Van~Noord \bgroup et al.\egroup
  }2013]{vannoord:2013}
Van~Noord, G., Bouma, G., Van~Eynde, F., de~Kok, D., van~der Linde, J.,
  Schuurman, I., Sang, E. T.~K., and Vandeghinste, V.
\newblock (2013).
\newblock Large scale syntactic annotation of written {D}utch: {L}assy.
\newblock In Peter Spyns et~al., editors, {\em Essential Speech and Language
  Technology for Dutch: Results by the STEVIN programme}, pages 147--164.
  Springer Berlin Heidelberg, Berlin, Heidelberg.

\bibitem[\protect\citename{Wijffels}2020]{wijffels:2020}
Wijffels, J., (2020).
\newblock {\em udpipe: Tokenization, Parts of Speech Tagging, Lemmatization and
  Dependency Parsing with the 'UDPipe' 'NLP' Toolkit}.
\newblock R package version 0.8.4-1.

\bibitem[\protect\citename{Wilcox}in press]{wilcox:2021}
Wilcox, R.~R.
\newblock (in press).
\newblock {\em Introduction to Robust Estimation and Hypothesis Testing}.
\newblock Academic press, San Diego, CA, 5th edition.

\bibitem[\protect\citename{Yarowsky \bgroup et al.\egroup }2001]{yarowsky:2001}
Yarowsky, D., Ngai, G., , and Wicentowski, R.
\newblock (2001).
\newblock Inducing multilingual text analysis tools via robust projection
  across aligned corpora.
\newblock In {\em Proceedings of the 1st International Conference on Human
  Language Technology Research (HLT)}, pages 161--168.

\end{thebibliography}

\end{document}